\documentclass[runningheads]{llncs}

\usepackage[T1]{fontenc}
\usepackage{graphicx}
\usepackage{tabularx}
\usepackage{longtable}
\usepackage{multirow}
\usepackage{booktabs,cellspace}
\usepackage{comment}
\usepackage{hyperref}
\usepackage{makecell}
\usepackage{wrapfig}
\usepackage[utf8]{inputenc}
\usepackage{color, colortbl}
\usepackage{csquotes}
\usepackage{pgfplots}

\usepackage{amsmath}
\usepackage{amssymb}
\usepackage{mathtools}

\usepackage{algorithm}
\usepackage{algpseudocode}
\usepackage{bm}
\usepackage{pifont}

\definecolor{ao(english)}{rgb}{0.0, 0.5, 0.0}
\usepackage{tikz}

\usepackage{siunitx}

\definecolor{cadmiumgreen}{rgb}{0.0, 0.42, 0.24}
\definecolor{cinnamon}{rgb}{0.82, 0.41, 0.12}
\definecolor{burntumber}{rgb}{0.54, 0.2, 0.14}
\definecolor{circlegreen}{RGB}{78,145,6}
\definecolor{circlered}{RGB}{164,0,0}

\newcommand{\mypar}[1]{\smallskip\noindent\textbf{#1.}}

\makeatletter
\newcommand{\printfnsymbol}[1]{%
  \textsuperscript{\@fnsymbol{#1}}%
}
\makeatother

\begin{document}

\sisetup{
  group-minimum-digits=4,   
  round-mode=places,        
  round-precision=2,       
  retain-explicit-plus,    
}

\title{Decision-Aware Suffix Prediction and Reasoning of Business Processes}

\titlerunning{Decision-Aware Suffix Prediction}

\author{Henryk Mustroph\orcidID{0009-0005-1946-1979}
\and Stefanie Rinderle-Ma\orcidID{0000-0001-5656-6108}}

\authorrunning{H. Mustroph and S. Rinderle-Ma}

\institute{Technical University of Munich, TUM School of Computation, Information and Technology,    Garching, Germany\\
\email{\{henryk.mustroph,stefanie.rinderle-ma\}@tum.de}}

\maketitle        

\begin{abstract}
Suffix prediction forecasts the remaining sequence of events of a running case until completion. Most approaches rely on neural networks trained on event logs, which, on average, perform well but struggle with short prefixes or targets belonging to a rare process variant. In such scenarios, the correct path may cross multiple branching decisions, determined primarily by case- and event-level attributes, a signal that NN-based suffix prediction models tend to underweight because they may heavily weight (dense) event labels. Decision mining extracts rules for such decisions from the event log, but has so far been applied only to post-hoc and what-if analysis, not suffix prediction. We therefore extend suffix prediction with decision mining, introducing a decision-aware suffix prediction framework, a neuro-symbolic approach that enables reasoning about predicted events via mined decision rules. Experiments on three of four event logs and three suffix predictors show that the framework can improve suffix prediction, especially for short prefixes but also for rare process variants, and adds intrinsic interpretability.

\keywords{Predictive Process Monitoring \and Suffix Prediction \and Decision Mining \and Neuro-Symbolic \and Interpretability}
\end{abstract}

\section{Introduction}\label{sec:intro}
Suffix prediction is a task in Predictive Process Monitoring (PPM) that forecasts the remaining sequence of events for a running case until completion. Suffix prediction enables proactive process interventions, such as early corrective actions and planning~\cite{Ceravolo}, e.g., Apromore applies PPM, including suffix prediction, to improve process transparency, reduce friction, and support resource planning\footnote{\tiny Apromore's PPM use cases:~\url{https://apromore.com/predictive-process-monitoring}}. Most suffix prediction approaches rely on neural networks (NNs)~\cite{camargo,evermann-suffix,psp,taymouri,Wuyts_sutran}. While early approaches trained only on event labels~\cite{evermann-suffix} or added temporal and resource features~\cite{camargo}, recent works show that using all case-level (i.e., static) and event-level (i.e., dynamic) attributes, i.e., payload data, as input yields the best predictive performance~\cite{Gunnarsson_ppm,psp,Wuyts_sutran}. This is intuitive, as payload data usually determines which branches are taken and whether events repeat in a trace. Suffix prediction becomes particularly difficult when given short prefixes or when the target suffix belongs to a rare process variant, since branches and loops are then highly probable. In these cases, NN-based suffix predictors must pay special attention to payload data. Yet payload data is often sparse in event logs, whereas event labels are mostly dense, an imbalance that leads to overemphasizing event-label patterns and underweighting decisive payload data, a behavior akin to shortcut learning~\cite{shortcut}.

Neuro-symbolic PPM~\cite{compl_neuro_symbol,neuro-symbol_ppm} integrates NNs with process-contextual symbolic knowledge to mitigate small, sparse training data and limited reasoning over process behavior. Existing approaches encode declarative control-flow knowledge to improve suffix prediction~\cite{neuro-symbol_ppm} or prescriptive compliance rules over control-flow, temporal, and other data conditions to constrain outcome prediction~\cite{compl_neuro_symbol}. Crucially, both rely on external rules and do not capture descriptive, data-derived knowledge, e.g., how payload data governs branching decisions. We therefore propose a neuro-symbolic suffix prediction framework that extracts symbolic knowledge directly from the event log via decision mining. Decision mining identifies the decision points (i.e., branches) of a process model, trains a decision tree per decision point on payload data, and extracts a decision rule for each branch~\cite{DM_deleonie,Rozinat_dm}. This has been applied to post-hoc (e.g., as part of process mining) or what-if analyses (e.g., business process simulation), but to our knowledge not to suffix prediction (i.e., as integration into PPM). Combining suffix prediction and decision mining yields two advantages: i) it lets the model better reflect the influence of payload data at branches, improving predictive performance, and ii) it enables intrinsic interpretation through conditional decision-rule-based justification of each predicted event after a branch returning the matched payload as the reason for the prediction. Our \emph{decision-aware suffix prediction framework} consists of a \emph{decision-aware event labeling} component, a \emph{decision-aware training}, and a \emph{decision-aware decoding and reasoning} procedure.

\textbf{Example:} In a procurement process, a requisition reaches a decision point after review and is either $\texttt{approved}$ or $\texttt{rejected}$ for revision. Since rejection and revision cycles are frequent in historical cases, a suffix predictor may overemphasize the event-label sequence pattern $\texttt{reject} \rightarrow \texttt{revise}$ while underusing decisive payload data, including static case-level and dynamic event-level attributes. Predicting another rejection can lead the entire predicted suffix down a revision path, even when the requisition already meets the approval conditions. However, after one revision cycle, the last event in the prefix has $\textit{amount} \ge 6000$ and $\textit{budget status} = approved$, indicating approval according to the discovered decision rule. A derived decision model captures this branching behavior, guides suffix prediction, and provides a decision-rule-based justification for the $\texttt{approve}$ prediction using these payload values. 

Experiments on one artificial and three real-life event logs, using three suffix predictors, demonstrate that the approach is model-agnostic, can improve suffix prediction performance and provides decision-rule-based justifications of predicted events at branches. We present our approach in Sect.~\ref{sec:method} and~\ref{sec:train-dec}, followed by the evaluation in Sect.~\ref{sec:eval}, related work in Sect.~\ref{sec:rel_work}, and the discussion and conclusion in Sect.~\ref{sec:concl}.
 
\section{Decision-Aware Suffix Prediction Framework}\label{sec:method}
\begin{figure}[htb!]
    \centering
    \includegraphics[width=0.70\linewidth]{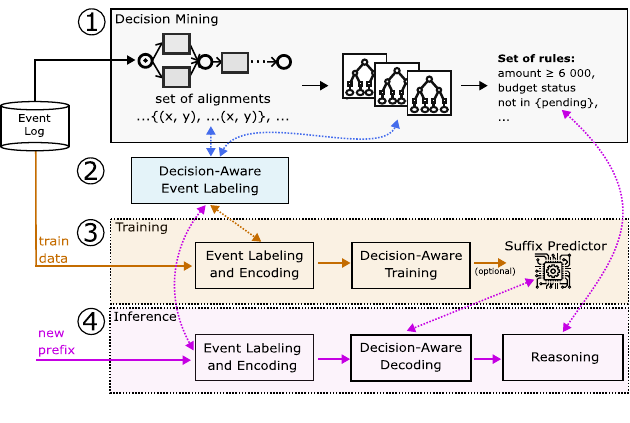}
    \caption{Decision-Aware Suffix Prediction Framework}
    \label{fig:approach}
\end{figure}

This section presents an overview of the proposed decision-aware suffix prediction framework, as illustrated in Fig.~\ref{fig:approach}. The framework consists of four components: decision mining \ding{192}, which builds the foundation for \emph{decision-aware event labeling} \ding{193} (presented in this section), which is required for \emph{decision-aware training} \ding{194} (presented in Sect.~\ref{sec:training}), and \emph{inference-time decision-aware decoding and reasoning} \ding{195} (presented in Sect.~\ref{sec:decoding}), which can be used together or independently.

Suffix prediction predicts the remaining sequence of events (i.e.,\ the suffix) of a running case, given its observed events (i.e.,\ the prefix). For a trace of length $M$, we define a prefix as $\pi_{\leq k} := \langle e_{1}, \dots, e_{k} \rangle$, and the corresponding suffix as $\sigma_{> k} := \langle e_{k+1}, \dots, e_{M}, \texttt{EOS} \rangle$,  where \texttt{EOS} denotes a special end-of-sequence event. Each event has an event label drawn from a finite alphabet $\mathcal{A}^{+}:= \mathcal{A} \cup \{\texttt{EOS}\}$, a timestamp, and may include multiple case-level (static) and event-level (dynamic) attributes (i.e., payload data).

\subsection{Decision Mining}
We adapt the decision mining approach of \cite{DM_deleonie} for our framework. The approach uses alignment-based conformance checking \cite{carmona_alignment} to replay traces on a process model, gathering, for each identified decision point, a dataset used to train a decision model. The discovered process model is represented as a Petri net \cite{DM_deleonie}. A Petri net is a tuple $N = (P, T, F)$, where $P$ is a set of places, $T$ is a set of transitions, and $F \subseteq (P \times T) \cup (T \times P)$ is the flow relation. For a transition $t \in T$, its preset and postset are defined as ${}^\bullet t := \{p \in P \mid (p,t) \in F\}$, and $t^\bullet   := \{p \in P \mid (t,p) \in F\}$. Each place in the Petri net $p \in P$ with multiple outgoing transitions ($|p^\bullet| > 1$) is called a decision point. For each decision point, we train a decision model with two objectives: first, it predicts the next event label after the decision with high accuracy, as its predictions are integrated into suffix prediction training and decoding; second, it provides an explanation, in the form of a decision rule, for why a particular event label is predicted, required for reasoning about the suffix prediction. \cite{DM_deleonie} was designed for post-hoc analysis and what-if analysis (e.g., data-driven simulation \cite{rozinat}) and relies on C4.5 decision trees, which support simple rule extraction by tree traversal, but often show limited predictive performance when decisions depend on complex data patterns. We therefore argue for a model architecture that better balances predictive performance and intrinsic interpretability. After discovering the process model, we replay each trace in the training set using alignment-based conformance checking \cite{carmona_alignment}, yielding a set of optimal alignments $\Gamma^{\mathrm{opt}}$. Based on the set, we construct decision-point datasets and train decision models.

\mypar{Decision Point Dataset and Model} 
An optimal alignment is defined as a sequence $\gamma = \langle (x_1, y_1), \dots, (x_n, y_n) \rangle \in \Gamma^{\mathrm{opt}}$, where $x_i$ denotes the move in the log (i.e., an event label or $\gg$) and $y_i$ denotes the move in the process model (i.e., a transition or $\gg$). We iterate over the set of alignments and create a dataset $I(p)$ for each decision point: for each position $i$ where $y_i = t \neq \gg$ and transition $t$ is preceded by a decision point $p \in {}^\bullet t$ with $|p^\bullet| > 1$, we add a supervised training instance $(\eta, a^*)$ to $I(p)$. The data state $\eta$ contains the case-level and event-level attribute values from the event observed at the most recent synchronous move before position $i$, denoted by $j:= \max \{ r < i \mid x_r \neq \gg \land y_r \neq \gg \}$, as well as the average event-level attribute values over all earlier synchronous moves $\{ r < j \mid x_r \neq \gg \land y_r \neq \gg \},$ to capture earlier values that may influence later decisions. The target label $a^* \in \mathcal{A}^{+}$ is the next visible event-label that follows $p$. Because the discovered Petri net contains silent transitions, $t$ may be a model move whose label is undefined (i.e., not existent in the log). Hence, $a^*$ cannot be read off from position $i$ alone. We therefore follow the token produced by firing $t$ forward along the alignment, through any subsequent model moves, to the first synchronous move $u:= \min \{ r \geq i \mid x_r \neq \gg \land y_r \neq \gg \land y_r \text{ consumes a token descended from the firing at } i \}$. Tracking token descent ensures that, under concurrency, $a^*$ is taken from the branch opened at $p$ rather than from an interleaved parallel branch. If such a synchronous move exists, we set $a^* = x_u$; if the branch reaches the final marking $m_f$ without any further synchronous move, we set $a^* = \texttt{EOS}$ to indicate that the decision at $p$ terminates the case. For each decision point, we train a classifier $\psi_p = \textsc{DecisionModel}(I(p))$ that predicts the probability distribution over next event labels and enables the extraction of rules for all possible next event labels. Unlike \cite{DM_deleonie}, which uses only case-level attributes as features and Petri net transitions as targets, our data states also include event-level attributes (i.e., aggregated historical and the most recent), with event labels as targets.

\subsection{Decision-Aware Event Labeling} \label{sec:labeling}
Building on the decision mining component, we define \emph{decision-aware event labeling}, applied offline before decision-aware training and online during inference-time decision-aware decoding. 

\mypar{Offline Labeling for Training}
Two training paradigms are commonly used for suffix prediction: next-event training, which optimizes the model to predict the next event \cite{camargo}, and sequence training, which predicts a sequence of events autoregressively \cite{Gunnarsson_ppm}, typically with teacher forcing, i.e., conditioning each step either on the ground-truth previous event or on the model's own previous prediction \cite{taymouri,psp}. Therefore, we label the full traces in the training set offline. We compute the optimal alignment for each trace and identify all synchronous moves whose move in the model (labeled transition) is followed by a decision point. The corresponding move in the log (event) $e_i$ is then called \emph{decision-labeled}, and we set $p_i := p$ and $z_i := \psi_{p_i}(\cdot \mid A_i) \in \Delta(\mathcal{A}^{+})$, where $A_i:=\eta_i$ is the data state up to and including $e_i$, and $z_i$ is the probability distribution over next-event labels returned by the trained decision model $\psi_{p_i}$. Otherwise, we set $p_i:= \bot$. For tractability, we adopt two simplifications, whose refinement is left for future work. First, if multiple decision points follow the same transition for a given event, we randomly select one, which may introduce noise. Second, we do not assign decision-aware labels to events generated during training, as this would require evaluating $\psi_{p_i}$ online for each sampled event in every batch across all epochs; ground-truth events used in teacher forcing are already decision-labeled, so the limitation applies only to the model's own predictions when used as input.

\mypar{Online Labeling for Decoding}
At inference time, only a prefix of a running case is available, so no offline alignment can be computed. Therefore, we construct decision-aware labels online by mapping each observed $e_i$ or generated event $\hat{e}_i$ to its corresponding transition $t$ in the Petri net and checking whether $t^\bullet$ contains a decision point $p$. The data state $\eta_i$ is constructed from the case-level attributes of the prefix and the event-level attributes observed or generated up to and including $\hat{e}_i$. If $e_i$ or $\hat{e}_i$ is decision-labeled, we store $p_i := p$, $A_i := \eta_i$, and $z_i := \psi_{p_i}(\cdot \mid A_i) \in \Delta(\mathcal{A}^{+})$, where $z_i$ is the probability distribution over next event labels predicted by the decision model $\psi_{p_i}$. We also store $A_i$ because the reasoning component uses it to explain decision-labeled event predictions during suffix generation.

\section{Decision-Aware Training and Decoding}\label{sec:train-dec}
This section presents the two decision-aware procedures of the framework, which can be applied together or independently.

\subsection{Decision-Aware Training}\label{sec:training}
We propose \emph{decision-aware training} for suffix prediction as a neuro-symbolic procedure that combines a base loss with a semantic loss \cite{semantic_loss_xu} that targets the next event label distribution for decision-labeled events.

\mypar{Base Loss}
The suffix predictor is trained to predict, for each event, its label and an optional set of event-level attributes $\mathcal{X}$ (e.g., resource and time features). The latter are included as prediction targets only when decision-aware decoding is applied and the decision models require them as inputs. The model input depends on the training paradigm, where $S^{(i)}$ denotes the number of decoding steps for training instance $i$. In next-event training, $S^{(i)}=1$: the model performs a single step ($s=0$) on the prefix. In teacher-forced sequence training, when $S^{(i)}>1$, the model decodes autoregressively: at $s=0$, the input is the prefix, and at each $s>0$, it is either the target suffix event or the event generated at step $s-1$. At each step $s\in\{0,\dots,S^{(i)}-1\}$, the model predicts an event label distribution
$\hat{p}_\theta^{(i,s)}(\cdot)\in\Delta(\mathcal{A}^{+})$ and, per optional event-level attribute $x\in\mathcal{X}$, either a distribution (categorical) or a scalar (continuous) $\hat{q}_\theta^{(i,s,x)}(\cdot)$. Each prediction target has a dedicated output head and all heads share the same input. The base loss sums over all heads:
\[
  \mathcal{L}_{\text{base}}
  \;=\;
  \mathcal{L}_{\text{label}}
  \;+\;
  \sum_{x\in\mathcal{X}}\lambda_x\,\mathcal{L}_{x},
\]
where $\mathcal{L}_{\text{label}}$ is the loss on the event label, each $\mathcal{L}_x$ is a loss specific to an event-level attribute $x$, and the weights $\lambda_x\ge 0$ balance the event-level attribute losses.

\mypar{Semantic Loss}
The semantic loss~\cite{semantic_loss_xu} connects the neural predictor to the decision model at each decision-labeled event.  Decision labeling for the training component is done offline (cf. Sect.~\ref{sec:labeling}). It penalizes the predictor for placing too little probability mass on labels the decision model considers plausible. Plausible labels are determined by thresholding $z$ at $\tau\in(0,1]$, such that those with probability $\geq\tau$ form the admissible set. For example, given $z = (a_1{:}\,0.9,\; a_2{:}\,0.08,\; a_3{:}\,0.02)$ over event labels $a_1,a_2,a_3$ and $\tau = 0.5$, only $a_1$ is admissible. Based on $\tau$, we define the admissible set of event labels, at decoder step $s$ with event $e^{(i)}_{k+s}$ (where $k$ denotes the prefix length of instance $i$), as $\mathcal{S}^{(i)}_{k+s} := \bigl\{a\in\mathcal{A}^{+}\;\big|\;z^{(i)}_{k+s}(a)\ge\tau\bigr\}$, and the per-step semantic loss is the negative log of the probability mass that $\hat{p}_\theta^{(i,s)}$ assigns to $\mathcal{S}^{(i)}_{k+s}$:
\[
  \ell_{\text{sem}}^{(i,s)}
  \;=\;
  \mathbf{1}\!\left[p^{(i)}_{k+s}\neq\bot\right]\,
  \mathbf{1}\!\left[\mathcal{S}^{(i)}_{k+s}\neq\emptyset\right]\,
  \mathbf{1}\!\left[a^{(i)}_{k+s+1}\in\mathcal{S}^{(i)}_{k+s}\right]\,
  \left(
    -\log\!\!\sum_{a\in\mathcal{S}^{(i)}_{k+s}}\!\!
        \hat{p}_\theta^{(i,s)}(a)
  \right).
\]
The loss is applied only at steps where the ground-truth next event $a^{(i)}_{k+s+1}$ falls within the admissible set $\mathcal{S}^{(i)}_{k+s}$, i.e., where the symbolic constraint agrees with the observed outcome; without this restriction, the semantic loss would penalize correct predictions whenever the decision model makes an incorrect one.

\mypar{Combined Loss}
For a mini-batch $\mathcal{B}$, the aggregate semantic loss and the final decision-aware objective are:
\[
  \mathcal{L}_{\text{sem}}
  \;=\;
  \frac{1}{N_{\mathcal{B}}}
  \sum_{i\in\mathcal{B}}\sum_{s=0}^{S^{(i)}-1}
    \ell_{\text{sem}}^{(i,s)},
  \qquad
  \mathcal{L}_{\text{DA}}
  \;=\;
  \mathcal{L}_{\text{base}}
  \;+\;
  \lambda_{\text{sem}}\,\mathcal{L}_{\text{sem}},
\]
with
\(
N_{\mathcal{B}}
\;=\;
\sum_{i\in\mathcal{B}}\sum_{s=0}^{S^{(i)}-1}
\mathbf{1}\!\left[p^{(i)}_{k+s}\neq\bot\right]\,
\mathbf{1}\!\left[\mathcal{S}^{(i)}_{k+s}\neq\emptyset\right]
\mathbf{1}\!\left[a^{(i)}_{k+s+1}\in\mathcal{S}^{(i)}_{k+s}\right]
\)
and regularization strength $\lambda_{\text{sem}}\ge 0$. For simplicity, we assume predicted event-level attributes to be independent of the predicted event label, although in practice they are not (e.g., the assigned resource depends on the performed activity). We leave this to future work.

\subsection{Decision-Aware Decoding and Reasoning at Inference}\label{sec:decoding}
At inference time, whenever the current decoding step corresponds to a decision point, \emph{decision-aware decoding and reasoning} combines the suffix predictor's label distribution with the decision model's distribution to improve predictive performance, and matches the payload of the decision-labeled event against the mined decision rules for the predicted label, yielding an interpretable justification. The idea is related to reachability-based masking for next-label prediction in~\cite{decode_masking}, where the prefix is replayed on the process model to obtain the current marking, and the predicted label distribution is hard-masked to reachable events only, renormalizing the remaining mass. Our method differs in three respects: the mask is derived from decision model probabilities rather than from global reachability alone, it is applied only locally at decision-labeled events rather than at every decoding step, and it reweights the suffix predictor's output instead of eliminating unsupported candidates, thereby preserving the learned distribution.

\mypar{Decoding}
Given a new prefix $\pi_{\le k}$, decoding proceeds for steps $r = 0, 1, \dots$ until \texttt{EOS} is predicted or a maximum length is reached. At each step, the model takes as input $e_{k+r}$, where $e_{k+0} = e_k$ is the last observed prefix event and $e_{k+r} = \hat{e}_{k+r}$ is the previously predicted event for $r > 0$, and outputs a distribution over $\mathcal{A}^{+}$ together with the event-level attribute predictions $\hat{q}_\theta^{(r,x)}$, $x\in\mathcal{X}$. The predicted event label $\hat{a}_{k+r+1}$ and the predicted event-level attributes jointly form the next event $\hat{e}_{k+r+1}$. At each step, we apply online decision labeling (Sect.~\ref{sec:labeling}). When $e_{k+r}$ is decision-labeled with tuple $(p_{k+r}, A_{k+r}, z_{k+r})$, we reweight the suffix predictor's next event label distribution $\hat{p}_\theta^{(r)}\in\Delta(\mathcal{A}^{+})$ by a decision-rule mask
\[
  m_r(a) := \bigl(\varepsilon + (1-\varepsilon)\,z_{k+r}(a)\bigr)^{\beta_r},
  \qquad a\in\mathcal{A}^{+},
\]
where $0 < \varepsilon \ll 1$ prevents exact-zero weights, and $\beta_r:= \beta_{\max}\exp(-\alpha r)$ controls the sharpness of the reweighting: $\beta_{\max}\ge 0$ sets the maximum guidance strength and $\alpha\ge 0$ is a decay rate that reduces guidance at later steps, where the data state is accumulated from predicted (and thus less
reliable) attributes. The decision-guided distribution is:
\[
  \tilde{p}_\theta^{(r)}(a)
  = \frac{\hat{p}_\theta^{(r)}(a)\,m_r(a)}
         {\sum_{b\in\mathcal{A}^{+}}\hat{p}_\theta^{(r)}(b)\,m_r(b)},
  \qquad a\in\mathcal{A}^{+}.
\]
As $\beta_r\to 0$, no decision guidance is provided; as $\beta_r\to\infty$, decoding collapses onto full decision guidance, defining a continuous family of symbolic constraints. If $e_{k+r}$ is not decision-labeled, $\tilde{p}_\theta^{(r)} := \hat{p}_\theta^{(r)}$. Three decoding strategies are supported for selecting $\hat{a}_{k+r+1}$. Mode (i.e., greedy) decoding (cf.~\cite{camargo}) selects the arg-max event label under $\tilde{p}_\theta^{(r)}$. Beam search (cf.~\cite{taymouri}) expands and scores candidate suffixes by accumulated log-likelihood under $\tilde{p}_\theta^{(r)}$, with each beam maintaining its own marking and data state. Stochastic decoding, e.g., Monte Carlo suffix sampling (MC-SA) (cf.~\cite{psp}), samples the next label from $\tilde{p}_\theta^{(r)}$.

\mypar{Reasoning}
At each decoding step where the current input event is decision-labeled, we produce a structured interpretation for the generated suffix. An interpretation is provided only when the next event label selected by the suffix predictor belongs to the data-supported branch set of the decision model at that decision point, i.e., $\{a\in\mathcal{A}^{+}: z_{k+r}(a) \geq \tau\}$ for support threshold $\tau$ (the same threshold as in the semantic loss of Sect.~\ref{sec:training}, although it may be set independently). We call such a step non-conflicting; a conflict is a decision-labeled step whose selected label falls outside this set. If so, we inspect the data state of the decision-labeled event, compare each attribute value against the decision-rule set prescribed for the predicted next event label, and return the tuple
\[
  \Bigl(
    p_{k+r},\;\;
    e_{k+r}\rightarrow\hat{a}_{k+r+1},\;\;
    \bigl\{(\texttt{attr},\,v,\,\mathrm{True}/\mathrm{False})\bigr\}
  \Bigr),
\]
where $p_{k+r}$ is the decision point, $e_{k+r}\rightarrow\hat{a}_{k+r+1}$ is the transition relation, and each triple $(\texttt{attr},\,v,\,\mathrm{True}/\mathrm{False})$ records an attribute name, its current value drawn from $A_{k+r}$, and whether that value belongs to the corresponding decision-rule set for $\hat{a}_{k+r+1}$.
\section{Evaluation}\label{sec:eval}
We evaluate the predictive performance and interpretability effects of the \emph{decision-aware suffix prediction framework} on three NN-based suffix predictors. Sect.~\ref{sec:eval:1} introduces the datasets and describes the experimental setup. Sect.~\ref{sec:eval:2} presents and discusses the results. The full implementation is available in our repository~\footnote{{\tiny Repository: \url{https://github.com/henryk-mustroph/decision_aware_suffix_prediction_framework}}}.

\subsection{Setting} \label{sec:eval:1}
We first introduce the datasets, the setup, and the configurations of all models.

\mypar{Datasets}
We evaluate on three real-life and one artificial dataset. The Helpdesk \footnote{{\tiny Helpdesk: \url{https://doi.org/10.4121/uuid:0c60edf1-6f83-4e75-9367-4c63b3e9d5bb}}} dataset is an event log from a ticket management system from an Italian software company. The Sepsis \footnote{{\tiny Sepsis:  \url{https://doi.org/10.4121/uuid:915d2bfb-7e84-49ad-a286-dc35f063a460}}} dataset represents the pathway of patients diagnosed with Sepsis through a hospital. The BPIC20 Domestic Declarations (BPIC20 DD)\footnote{\tiny BPIC20 DD: \url{https://data.4tu.nl/datasets/6a0a26d2-82d0-4018-b1cd-89afb0e8627f}} dataset is a travel expense claims process and contains events over two years. The artificial Procurement dataset is generated by a custom Python simulator (available in the repository) and contains 16 attributes per event, three XOR decision points (one loop) governed by deterministic decision rules, and variable-length traces of 10--15 events.

\mypar{Decision Mining}
First, we derive a process model from the event log using the Inductive Miner (IM)~\cite{in_miner} from PM4Py\footnote{{\tiny Inductive Miner: \url{https://processintelligence.solutions/pm4py/features}}}, which returns a process tree that is converted into a Petri net. Any other discovery algorithm could be used as well. We chose the IM because it discovers sound, block-structured process models with well-defined decision points, in which each event label maps to exactly one non-silent transition. However, the IM guarantees a fitness of $1.0$, which risks overgeneralization (i.e., allowing more behavior than observed in the log). We therefore control the trade-off between fitness and precision via the IM noise threshold, setting it to $0.0$ for Helpdesk and Procurement (i.e., the logs are clean and structured), $0.15$ for BPIC20 DD, and $0.20$ for Sepsis (i.e., the logs lead to flower-like process models). Next, we applied alignment-based conformance checking, which is non-deterministic because multiple optimal alignments with the same minimal cost can occur, leading to different decision-point training data. Replaying traces multiple times showed only slight differences in the resulting decision-point datasets, so we assume a single replay provides adequate results. For all datasets, the data states are constructed uniformly as described in Sect.~\ref{sec:method}, without any dataset-specific feature engineering.

Our framework requires decision models that are both accurate and interpretable (i.e., discover meaningful decision rules). We therefore train two classifiers per decision point: a CatBoost\footnote{\tiny CatBoost: \url{https://catboost.ai/docs/en/concepts/python-reference_catboostclassifier}} model, which is a gradient-boosted decision tree, for accurate event label prediction,  and a decision tree (DT)\footnote{\tiny DecisionTree: \url{https://scikit-learn.org/stable/modules/generated/sklearn.tree.DecisionTreeClassifier.html}} as a surrogate model for decision rule extraction via tree traversal. In the event logs used, payload data is mainly categorical, sparse, and skewed, especially event-level features (e.g., resources). We therefore decided to use CatBoost, since it outperforms a standard DT on this kind of data. CatBoost ensembles shallow trees with target-statistic encoding instead of one-hot splits, and adds prior smoothing regularization for skewed features whose classes appear only a few times. Note that the benefit of the decision models over the NN suffix predictor does not stem from CatBoost itself: each decision model is a local classifier trained only on the aligned data states at its decision point, which isolates the payload data-to-decision point signal from the global sequence patterns that dominate suffix predictor training. However, other logs contain more numerical data, e.g., sensor data in manufacturing, for which a different tree-based model might be more suitable than CatBoost. Additionally, as an ensemble of trees, CatBoost yields no single traversable structure and thus no decision rules; understanding feature contributions would require XAI methods like SHAP, as in \cite{Park_dm}. We therefore pair CatBoost with a DT as its surrogate model to discover decision rules. A limitation is that a DT cannot always match CatBoost's predictions, so some predictions lack extractable rules. If interpretability is the priority, we recommend using the DT for both. CatBoost is implemented with a depth of $4$, $L_2{=}8.0$, and trained with a learning rate of $0.05$ for $300$ boosting iterations. The decision tree has a depth of $\leq 4$ and uses inverse-frequency class weights so that its rules surface rare-but-real minority branches rather than only the majority. Both require $\ge50$ decision-point observations per leaf.

\mypar{Suffix Predictors}
We implemented three NN-based suffix predictors: the full-shared LSTM (\emph{FS-LSTM}) with \emph{mode} decoding \cite{camargo}, the GAN-based encoder-decoder LSTM (\emph{GAN-LSTM}) with \emph{beam search} decoding \cite{taymouri}, and the uncertainty-aware encoder-decoder LSTM (\emph{U-ED-LSTM}) with \emph{MC-SA} decoding \cite{psp}. The U-ED-LSTM is the most recent of the three and, like SuTraN~\cite{Wuyts_sutran}, fully data-aware. In all models, categorical attributes are embedded, numerical features are standardized before concatenation, and categorical outputs use softmax heads. All models take the event labels and event-level attributes as input and, for each generated event, predict the event label and all dynamic features required by the decision models. Additionally, the U-ED-LSTM takes all case-level attributes as input. The \emph{FS-LSTM} passes the input through a shared LSTM, followed by a dedicated LSTM and dense layer per output feature (single layers, hidden size $50$). It is not autoregressive ($S{=}1$, i.e., next-event training), uses standard cross-entropy on the next event label as $\mathcal{L}_{\text{base}}$, and is trained with Adam, a learning rate of $10^{-3}$, and dropout of $0.1$. During inference, it rolls out the most likely suffix. The \emph{GAN-LSTM} generator is an encoder-decoder LSTM (five layers, hidden size $200$) whose generated suffixes are judged by an LSTM discriminator trained to distinguish real from generated suffixes. $\mathcal{L}_{\text{base}}$ combines the adversarial generator loss with standard supervised terms \cite{taymouri}. It is autoregressive ($S{=}5$) with scheduled teacher forcing starting at a ratio of $1.0$, trained with RMSprop, a learning rate of $5{\times}10^{-5}$, and dropout of $0.2$. During inference, it predicts $B{=}3$ suffix beams. The \emph{U-ED-LSTM} captures epistemic uncertainty via Monte Carlo dropout and aleatoric uncertainty via learned means and variances (encoder and decoder with four LSTM layers, hidden size $128$). $\mathcal{L}_{\text{base}}$ contains loss-attenuated supervised terms \cite{psp}. It is autoregressive ($S{=}5$) with scheduled teacher forcing starting at a ratio of $1.0$, trained with Adam, a learning rate of $10^{-5}$, and dropout of $0.1$. During inference, dropout stays active, and $T{=}100$ suffix samples are generated.

\mypar{Data Preparation, Model Features, and Hyperparameters}
We split the datasets at the case level into training, validation, and test sets at a $65\%$--$15\%$--$20\%$ ratio. We cut each trace into prefix--suffix pairs and append $S{=}5$ \texttt{EOS} events to every case. To enable batching, all prefixes are left zero-padded to a common window $W=\lceil q_{0.985}(\text{case lengths})\rceil + S$, categorical event attributes are index-encoded with index $0$ reserved for the left padding, and continuous attributes are $z$-standardized. The decision models take the same event-level attributes as the suffix predictors, plus all case-level attributes, as input. The decision-aware training hyperparameters are $\tau=0.2$, $\lambda_{\text{sem}}=0.3$ (GAN-LSTM, U-ED-LSTM) and $0.5$ (FS-LSTM), and $\lambda_{x}=1.0$. For the decision-aware decoding, we set $\varepsilon=10^{-3}$, $\beta_{\max}=2.0$, $\alpha=0.1$, and for the reasoning $\tau=0.2$. All values are determined via grid search, selecting the setting with the best validation performance. As a practical matter, the decoding parameters were transferred unchanged across all four logs, whereas $\lambda_{\text{sem}}$ required per-model tuning. End users should thus start from these defaults and tune $\lambda_{\text{sem}}$ and $\beta_{\max}$ on a validation split.

\subsection{Results}\label{sec:eval:2}
We first report the predictive performance of our framework and then present the
conditional decision-rule-based justification results that add intrinsic interpretability to the NN-based suffix predictors.

\mypar{Predictive Performance}
We evaluate suffix prediction performance using the normalized Damerau-Levenshtein similarity (DLS), a commonly used metric for suffix prediction:
\begin{equation}
    \mathrm{DLS}(\hat{\sigma},\sigma)
    := 1 - \frac{\mathrm{DL}(\hat{\sigma},\sigma)}
    {\max(|\hat{\sigma}|,|\sigma|)},
    \label{eq:dls}
\end{equation}
where $\sigma$ and $\hat{\sigma}$ are the ground-truth and predicted suffixes, respectively, and $\mathrm{DL}$ denotes the Damerau-Levenshtein edit distance. A score of $1$ indicates identical suffixes, whereas a score of $0$ indicates maximum dissimilarity. For \emph{mode} decoding, we compare the single argmax suffix with the ground truth. For \emph{beam search}, we compare the highest-ranked beam according to its cumulative log-probability with the ground truth. For \emph{MC-SA}, we compute the DLS for every sampled suffix and report the mean.

We compare the clean (standard) configuration, using the training and decoding procedures described in the respective original papers, with three decision-aware configurations: (i) decision-aware training with standard decoding, (ii) standard training with decision-aware decoding, and (iii) decision-aware training and decision-aware decoding. Table~\ref{tab:dls_main} reports the results for all three suffix predictors on all four datasets.
\begin{table}[ht]
\caption{DLS scores (\(\in [0,1]\) $\uparrow$) for each suffix predictor and all datasets. Differences to \emph{clean} are given in parentheses, positive differences in bold.}
\label{tab:dls_main}
\centering

\tiny

\setlength{\tabcolsep}{3pt}
\renewcommand{\arraystretch}{1.0}
\begin{tabular}{llcccc}
\toprule
\textbf{Model} & \textbf{Variant} & \textbf{Helpdesk} & \textbf{Sepsis} & \textbf{BPIC20 DD} & \textbf{Procurement} \\
\midrule

\multirow{4}{*}{FS-LSTM \cite{camargo}
}
& clean               & 0.644 & 0.327 & 0.889 & 0.860 \\
& decision train      & 0.635 (-0.009) & 0.325 (-0.002) & 0.888 (-0.001) & 0.854 (-0.006) \\
& decision decoding   & 0.866 (\textbf{+0.222}) & 0.129 (-0.198) & 0.931 (\textbf{+0.042}) & 0.885 (\textbf{+0.025}) \\
& dec. train + dec.   & 0.864 (\textbf{+0.220}) & 0.130 (-0.197) & 0.931 (\textbf{+0.042}) & 0.877 (\textbf{+0.017}) \\
\midrule

\multirow{4}{*}{GAN-LSTM \cite{taymouri}
}
& clean               & 0.868 & 0.330 & 0.925 & 0.896 \\
& decision train      & 0.868 (\textbf{+0.001}) & 0.316 (-0.014) & 0.928 (\textbf{+0.003}) & 0.899 (\textbf{+0.003}) \\
& decision decoding   & 0.871 (\textbf{+0.004}) & 0.086 (-0.244) & 0.931 (\textbf{+0.006}) & 0.913 (\textbf{+0.017}) \\
& dec. train + dec.   & 0.871 (\textbf{+0.004}) & 0.104 (-0.226) & 0.927 (\textbf{+0.002}) & 0.914 (\textbf{+0.018}) \\
\midrule

\multirow{4}{*}{U-ED-LSTM \cite{psp}
}
& clean               & 0.827 & 0.239 & 0.774 & 0.837 \\
& decision train      & 0.834 (\textbf{+0.007}) & 0.234 (-0.004) & 0.776 (\textbf{+0.002}) & 0.841 (\textbf{+0.004}) \\
& decision decoding   & 0.864 (\textbf{+0.037}) & 0.150 (-0.088) & 0.840 (\textbf{+0.067}) & 0.851 (\textbf{+0.014}) \\
& dec. train + dec.   & 0.867 (\textbf{+0.039}) & 0.149 (-0.089) & 0.840 (\textbf{+0.066}) & 0.851 (\textbf{+0.014}) \\
\bottomrule
\end{tabular}
\end{table}

Overall, decision-aware decoding produces the clearest improvements, whereas decision-aware training has only a small effect. The main exception is Sepsis, for which the decision-aware setting reduces predictive performance. One possible explanation is that the Sepsis log is highly variable: many test cases belong to variants not observed in the training data. Additionally, the cases often contain long, variable loops of a single event (self-loops) that depend on event-level patient measurements, many of which are continuous (e.g., leukocyte measurements) and difficult for the suffix predictor to predict accurately. These predictions are then used by the decision models, leading to higher error chains. Plain suffix prediction, in contrast, likely just predicts the average number of loop repetitions, giving it better overall DLS scores, though these are also only around $30\%$.

(i) \emph{Standard vs. decision-aware training.} Across model architectures and datasets, most changes in DLS are below $1\%$. Given that each dataset contains thousands of evaluated prefixes, the practical
effect of decision-aware training is therefore negligible. This result suggests that, in its current form, the auxiliary semantic loss $\mathcal{L}_{\mathrm{sem}}$ has too little influence on suffix prediction performance. The effect is slightly negative for the FS-LSTM and zero or slightly positive for the GAN-LSTM and U-ED-LSTM. This difference may be related to the FS-LSTM training configuration ($S=1$, without teacher forcing). The semantic loss adds a second training objective that slightly changes the next-event-label probabilities. The FS-LSTM is trained for single-step, next-event-label prediction, but is applied here for sequence prediction. The last predicted event is appended to the current sequence to predict the next event. As a result, small changes in the probabilities can accumulate more heavily across prediction steps than in the other two (sequence-trained) models. Future work should therefore investigate stronger ways of integrating decision knowledge during training. The soft auxiliary loss commonly used in neuro-symbolic learning appears too weak in this setting.

(ii) \emph{Standard vs. decision-aware decoding.} In contrast, decision-aware decoding yields clear improvements for every model on all datasets except Sepsis. The largest increase is $+0.22$ DLS for the FS-LSTM on Helpdesk. The other two predictors also improve, showing that decision-aware decoding is useful across different model architectures. In particular, the U-ED-LSTM already uses all payload attributes during training and decoding, yet still benefits from the decision-aware decoder. This indicates that the decision-aware decoder setting can add value even when the underlying suffix predictor is fully data-aware.

(iii) \emph{Standard vs. decision-aware training and decoding.} Because the effect of decision-aware training is negligible and the effect of decision-aware decoding is positive, the combined configuration is also positive overall and is dominated by the decoding effect. We find no evidence that the two components interfere with each other.

\begin{figure}[htb!]
    \centering
    \includegraphics[width=0.72\linewidth]{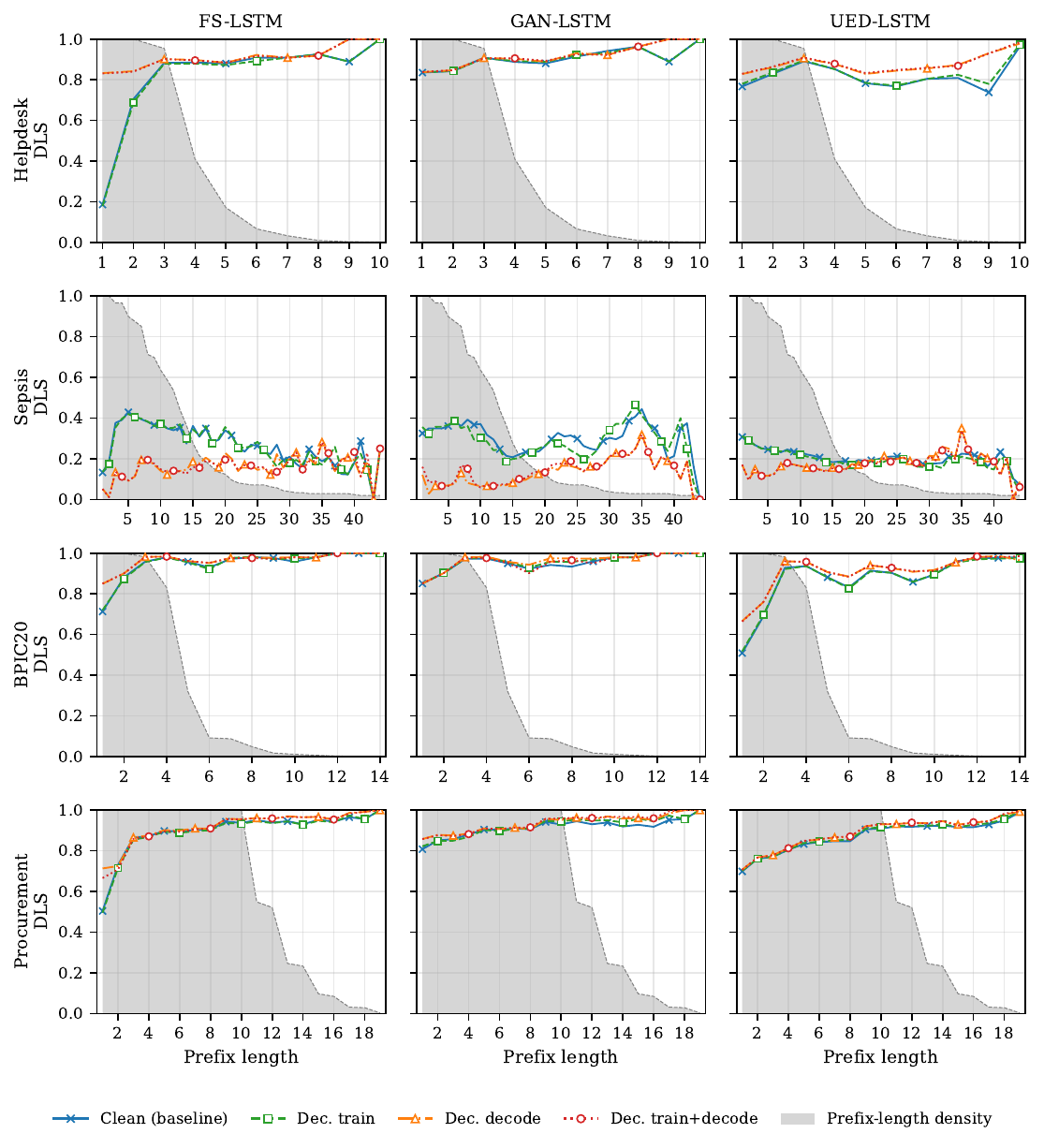}
    \caption{DLS of all suffix predictors and decoding configurations by prefix length.}
    \label{fig:dls_res}
\end{figure}

\begin{figure}[htb!]
    \centering
    \includegraphics[width=0.70\linewidth]{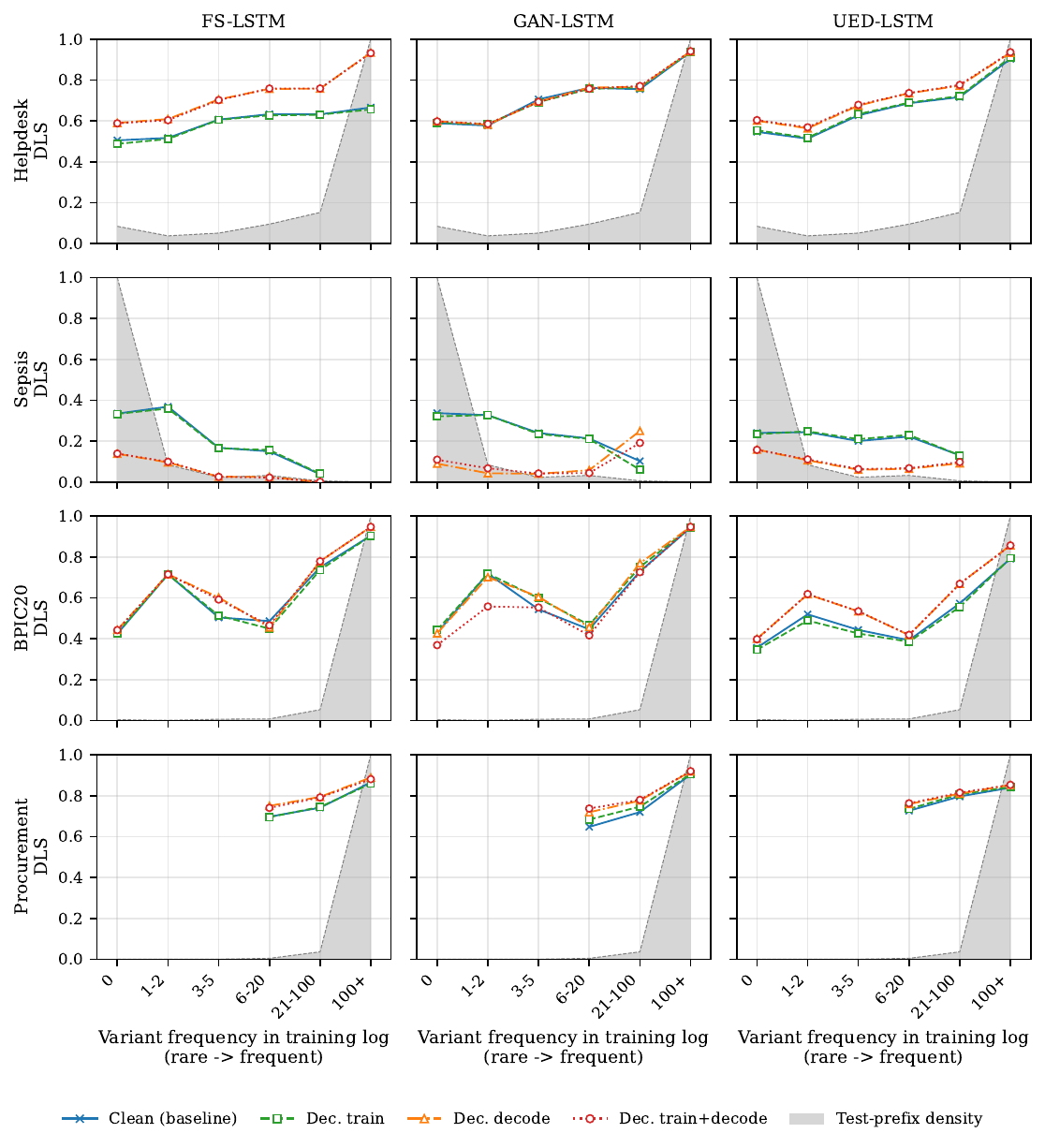}
    \caption{DLS of all suffix predictors and decoding configurations by the frequency of process variants.}
    \label{fig:dls_rarity}
\end{figure}

\mypar{Performance for Short Prefixes and on Rare Process Variants}
Both figures contain gray areas that indicate the density of prefixes with a specified length (Figure~\ref{fig:dls_res}) or the density of prefixes binned by frequency (rarity) (Figure~\ref{fig:dls_rarity}) in the test set.

Figure~\ref{fig:dls_res} reports the DLS scores for each model by prefix length. Decision-aware decoding
is particularly beneficial for short prefixes, which is consistent with the motivation for our approach: when only a small part of a case has been observed, the suffix predictor has little sequential context and can benefit from the additional decision information. The effect is strongest for the FS-LSTM, which
also explains why this model obtains the largest overall gain in Table~\ref{tab:dls_main}. The GAN-LSTM and U-ED-LSTM show similar improvements, even though only the U-ED-LSTM is fully data-aware and the GAN-LSTM is not. Because the GAN-LSTM and U-ED-LSTM improve to a similar extent, this comparison does not reveal a clear additional effect of providing all payload data to the suffix predictor itself.

Figure~\ref{fig:dls_rarity} also supports our second claim that decision-aware decoding improves suffix prediction for rare process variants. A variant is a test case's full event sequence. We group test prefixes by how often their variant occurs in the training and validation data, from $0$ (unseen) to $100+$. These frequencies are used only after inference, avoiding data leakage. On Helpdesk and BPIC20, DLS increases with variant frequency across all configurations. Since rare variants also tend to be longer, we compare rare ($\leq 5$ occurrences) and frequent variants of similar length. In the clean setting, rare variants still score $0.09$--$0.16$ DLS lower on Helpdesk and $0.20$--$0.26$ lower on BPIC20, showing that purely NN-based predictors struggle with rare variants. Decision-aware decoding improves DLS for rare variants by $0.09$ (FS-LSTM) and $0.05$ (U-ED-LSTM) on Helpdesk, and by $0.06$ and $0.07$, respectively, on BPIC20. The effect on GAN-LSTM is negligible. These gains are not consistently larger than for frequent variants. U-ED-LSTM benefits more on rare variants, whereas FS-LSTM benefits more on frequent variants in Helpdesk. Procurement and Sepsis do not support this comparison. In synthetic Procurement, every test variant occurs more than five times. In Sepsis, $74\%$ of test cases belong to unseen variants, leaving too few frequent variants for a fair comparison. This high share of unseen variants also helps explain why Sepsis is difficult to predict.

\begin{figure}[htb!]
    \centering
    \includegraphics[width=0.70\linewidth]{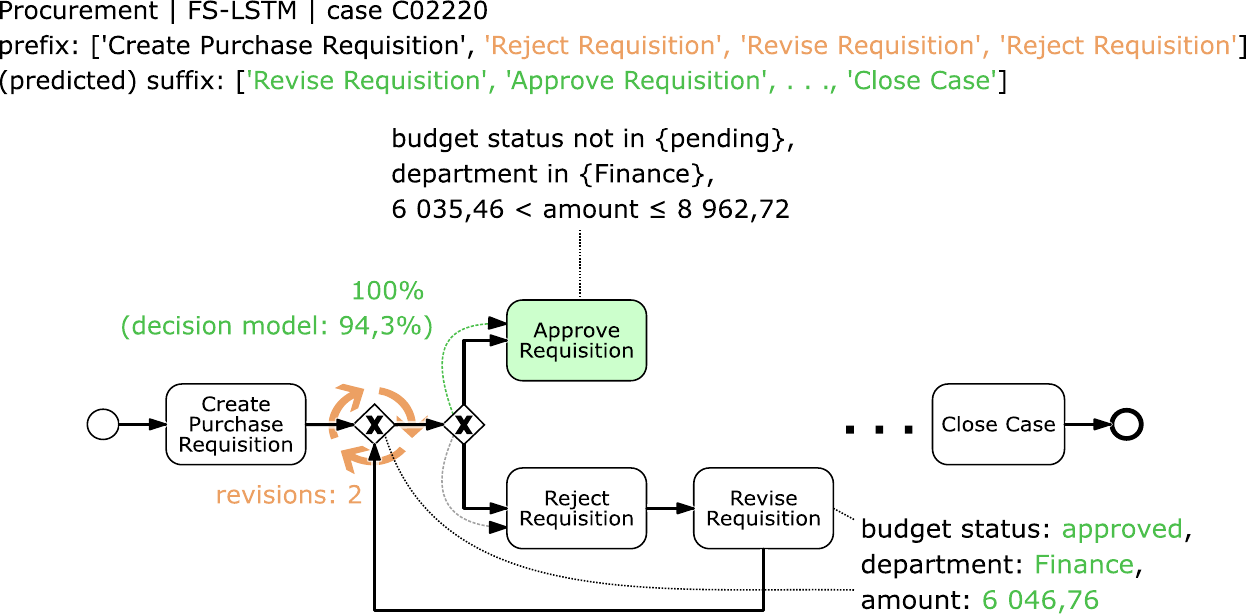}
    \caption{Reasoning capability and visualization for Procurement | C02220.}
    \label{fig:example}
\end{figure}

\mypar{Conditional Decision-Rule-Based Justification}
We modeled a (sub-)process of the artificial procurement event log in BPMN to demonstrate the reasoning capability of our approach, as shown in Fig.~\ref{fig:example}. We visualize the initial process segment, which contains the only looping and XOR decision, with the prefix and predicted suffix for case C02220 (FS-LSTM with standard training and decision-aware decoding) mapped onto it, together with the decision rule discovered for the $\texttt{approve}$ event and the predicted payload values of the $\texttt{revise}$ event that provide the reasoning for the $\texttt{approve}$ prediction. After two $\texttt{reject}$ events, the predicted $\texttt{revise}$ attribute values, $\textit{department} = \texttt{Finance}$ (case-level: given from the prefix), $\textit{budget status} = \texttt{approved}$ (event-level: predicted), and $\textit{amount} = 6{,}046.76$ (event-level: predicted), satisfy the decision rule for the $\texttt{approve}$ event, which is also the top next event label of the CatBoost model and predicted by the FS-LSTM with decision-aware decoding. Since these values match the decision rule of $\texttt{approve}$, they are indicated to the user as a reason for the prediction.

Additionally, we assess the conditional decision-rule-based justification component quantitatively via the interpretability rate $ir$, which is the fraction of non-conflicting decision-labeled steps in which all attribute conditions of the applicable decision rule are satisfied. $ir{=}1$ indicates that every such step is fully supported. In contrast, standard suffix prediction provides no justification at all, so every fully supported step adds interpretability to an otherwise black-box forecast. Across all models and decision-aware configurations, $ir$ is high on Helpdesk ($0.55$--$0.92$) and Sepsis ($0.71$--$0.93$), but markedly lower on BPIC20 DD ($0.17$--$0.37$) and Procurement ($0.26$--$0.48$). These rates must be read together with coverage, i.e., the share of decision-labeled steps for which an interpretation is attempted at all: coverage is large on Sepsis and Procurement but small on Helpdesk, so the high Helpdesk rates are computed over a few steps. The key result is thus mixed: in two logs, the majority of non-conflicting decision-labeled predictions receive a full rule-based justification, whereas in the other two logs, most predictions remain unexplained. Compared to the non-interpretable baseline, this is a net gain, but a user cannot yet rely on receiving an explanation for every decision-labeled prediction. We attribute the low rates to two factors. First, the surrogate decision tree approximates CatBoost imperfectly, so extracted rules can be missing or too coarse for the labels CatBoost predicts. Second, rule conditions are evaluated on predicted event-level attribute values, and a single mispredicted attribute breaks a conjunctive rule, which is most visible on Procurement, whose rules are deterministic conjunctions over multiple attributes. Notably, interpretability and predictive benefit behave independently. On Sepsis, $ir$ is high while decision-aware decoding degrades DLS, and on BPIC20 DD, DLS improves despite low $ir$, consistent with decoding relying on the CatBoost distribution directly rather than the surrogate DT.

\section{Related Work}\label{sec:rel_work}

\mypar{Decision Mining}
In \cite{Rozinat_dm}, decision mining was introduced by identifying decision points via token-based replay and extracting decision rules using simple decision models. \cite{DM_deleonie} replaced token-based replay with alignments, which yield more precise results, since even traces that deviate from the discovered process model can still be replayed and used. \cite{Mannhardt_dm} further improved decision mining by uncovering overlapping decision rules. The approaches \cite{DM_deleonie,Mannhardt_dm,Rozinat_dm} are descriptive, i.e., they are used post hoc to explain observed routing decisions. \cite{Scheibel_online_dm} proposed an online decision-mining approach that updates decision rules during process execution as relevant data becomes available over time. More recently, \cite{Park_dm} used NNs to improve next event-label prediction at decision points. Unlike traditional decision mining, their approach does not extract explicit decision rules but derives feature attributions via XAI methods, such as SHAP. However, they do not integrate decision mining into event sequence prediction.

\mypar{Process-Semantic Constrained PPM}
Recent work integrates explicit process knowledge into NN-based PPM. \cite{decode_masking} derives reachable events from a discovered Petri net to hard-constrain next-event label prediction during decoding. Neuro-symbolic approaches enforce process rules during training: \cite{neuro-symbol_ppm} compiles declarative control-flow constraints as LTL formulas into the loss, and \cite{compl_neuro_symbol} trains the model with Logic Tensor Networks to satisfy prescriptive compliance rules that link control-flow, temporal, and other data conditions to the prediction target. Both rely on externally specified, normative rules: \cite{neuro-symbol_ppm} uses no payload data as symbolic input and provides no interpretability for suffix prediction, while \cite{compl_neuro_symbol} targets outcome prediction and can override correct but non-compliant predictions. In contrast, our framework derives its symbolic knowledge from the event log via Process Mining, more precisely Decision Mining (i.e., descriptive).

\section{Discussion and Conclusion}\label{sec:concl}
This work integrates decision mining into suffix prediction via a decision-aware event-labeling strategy that enables decision-aware training, decision-aware decoding, and reasoning during inference. Decision-aware training has only a marginal effect on predictive performance, whereas decision-aware decoding consistently improves performance, especially for shorter prefixes.

\mypar{Benefit and Practical Applicability}
First, suffix prediction creates operational value by enabling timely intervention and planning. This is most effective early, when only a short prefix is observed, and the running case can still be steered. We could show that applying the decision-aware suffix prediction framework on standard models improves accuracy exactly there. Second, the adoption of suffix prediction in practice also depends on trust. A standard NN-based suffix prediction is hard for a domain expert to act on, and even harder to be accountable for \cite{Ceravolo}. The conditional decision-rule-based justification attaches a human-readable interpretability to each prediction along a branch, i.e., a decision-labeled event if the payload data matches the rule, thereby providing intrinsic interpretability to NN-based suffix predictors. Third, applying the framework is straightforward. All symbolic components (i.e., decision mining and labeling) run offline, require no manually specified rules or domain knowledge, and inference adds only a marking update and a decision model prediction per decision-labeled event. However, as a challenge here, the (event-level) payload data that actually drives decisions must be predictable along the suffix, otherwise, the decision models can neither guide nor explain the prediction.

\mypar{Limitation and Future Work}
First, the framework inherits the quality of its symbolic inputs: an imprecise process model yields spurious decision points, and weak decision models can misguide the decoding process. The decay parameter $\beta_r$ mitigates but does not eliminate this risk. As the Sepsis results show, on logs dominated by single-event loops and hard-to-predict continuous event-level attributes (i.e., patient measurements), decision guidance can even degrade predictive performance, so the framework should not be applied there. Second, the interpretability rate remains limited, as the surrogate DT approximates CatBoost imperfectly. Here, we plan to explore decision models that better balance predictive performance and interpretability (cf.\ Sect.~\ref{sec:eval:2}) in future work. Third, hyperparameters were selected via grid search, and we report no multi-seed runs, significance tests, or systematic sensitivity analysis. Since the decoding results are good and consistent across three architectures and three of the four event logs, we plan, for future work, to perform a more detailed ablation study on multiple seeds and on additional datasets.

\bibliographystyle{splncs04}
\bibliography{references}

\end{document}